\documentclass[10pt, a4paper]{article}

\usepackage[final]{lrec2026} 

\usepackage{hyperref}

\usepackage{xcolor}
\usepackage{xspace}
\usepackage{multirow}
\usepackage{graphicx}

\usepackage{enumitem}
\usepackage{amsmath}
\usepackage{booktabs}
\usepackage{hyperref}

\setlist{leftmargin=*,nosep}

\title{Structured Legal Document Generation in India: A Model-Agnostic Wrapper Approach with \texttt{VidhikDastaavej}}

\name{
\parbox{\textwidth}{
\centering
Shubham Kumar Nigam$^{1,2*\dagger}$ \quad
Balaramamahanthi Deepak Patnaik$^{1*}$ \quad \\
Noel Shallum$^{3}$ \quad
Kripabandhu Ghosh$^{4}$ \quad
Arnab Bhattacharya$^{1}$
}
}

\address{
$^{1}$Indian Institute of Technology Kanpur, India;
$^{2}$University of Birmingham Dubai, UAE \\
$^{3}$Symbiosis Law School Pune, India;
$^{4}$IISER Kolkata, India \\
\{shubhamkumarnigam, bdeepakpatnaik2002, noelshallum\}@gmail.com\\
kripaghosh@iiserkol.ac.in, \quad arnabb@cse.iitk.ac.in
}

\begin{document}

\abstract{
Automating legal document drafting can improve efficiency and reduce the burden of manual legal work. Yet, the structured generation of private legal documents remains underexplored, particularly in the Indian context, due to the scarcity of public datasets and the complexity of adapting models for long-form legal drafting.  
To address this gap, we introduce \texttt{VidhikDastaavej}, a large-scale, anonymized dataset of private legal documents curated in collaboration with an Indian law firm. Covering 133 diverse categories, this dataset is the first resource of its kind and provides a foundation for research in structured legal text generation and Legal AI more broadly.  
We further propose a Model-Agnostic Wrapper (MAW), a two-stage generation framework that first plans the section structure of a legal draft and then generates each section with retrieval-based prompts. MAW is independent of any specific LLM, making it adaptable across both open- and closed-source models.  
Comprehensive evaluation, including lexical, semantic, LLM-based, and expert-driven assessments with inter-annotator agreement, shows that the wrapper substantially improves factual accuracy, coherence, and completeness compared to fine-tuned baselines. This work establishes both a new benchmark dataset and a generalizable generation framework, paving the way for future research in AI-assisted legal drafting.  
\\ \newline  }

\maketitleabstract

\renewcommand{\thefootnote}{$*$}
\footnotetext{These authors contributed equally to this work}
\renewcommand{\thefootnote}{$\dagger$}
\footnotetext{Corresponding author}

\section{Introduction}
Automating legal document generation can significantly improve efficiency and accessibility in legal workflows. Although LLMs have been widely used for legal tasks such as prediction of judgments, summarization of cases, and retrieval, their application to the generation of private legal documents remains underexplored, particularly in the Indian legal domain. The primary challenge lies in the confidentiality of private legal documents, which limits publicly available training data.

To address this, we introduce {VidhikDastaavej}, a novel anonymized dataset of private legal documents, collected in collaboration with Indian legal firms. The name \texttt{VidhikDastaavej} is derived from the Hindi words ``Vidhik'' (legal) and ``Dastaavej'' (documents), reflecting its focus on legal document automation. This dataset serves as a valuable resource for training and evaluating structured legal text generation models, while ensuring compliance with ethical and privacy standards.

To further complicate matters, the landscape of large language models is evolving at a rapid pace, with new models being released frequently. In such a scenario, methods that rely on task-specific supervised fine-tuning (SFT) quickly become outdated or impractical, especially when a newer, more powerful model is introduced shortly after. Moreover, most end-users, such as legal practitioners or developers working with proprietary or custom-deployed models, may not have the resources to retrain or fine-tune large models. In some cases, users may prefer to keep their model private or operate within hardware constraints that prevent full-scale training. This raises an urgent need for model-agnostic approaches that can adapt seamlessly across different LLMs without requiring architectural modifications or extensive retraining.
To overcome this challenge, we propose a lightweight and scalable \textit{Model-Agnostic Wrapper (MAW)} for structured legal document generation. The wrapper decouples the generation process from any particular model by adopting a two-stage workflow: first generating section titles from document instructions, followed by iterative content generation for each section. This structure-then-generate strategy promotes coherence, reduces hallucinations, and ensures factual alignment, all while remaining compatible with any base LLM, whether open-source, commercial, or privately hosted. This flexibility makes our approach particularly valuable for real-world legal applications where model diversity and resource constraints are the norm.

For rigorous evaluation, we introduce expert-based assessment, where legal professionals review generated documents based on factual accuracy (adherence to legal instructions) and completeness and comprehensiveness (coverage of all essential details) between 1--10 (Irrelevant--Relevant) Likert scale. This ensures a robust evaluation beyond standard lexical and semantic metrics, addressing the complexity of legal drafting.

Additionally, we provide an interactive Human-in-the-Loop (HITL) Document Generation System, enabling users to input document types, customize sections, and generate structured legal drafts. To enhance reproducibility, we have made the \texttt{VidhikDastaavej} dataset, models, and codes available through a GitHub repository\footnote{\url{https://github.com/ShubhamKumarNigam/VidhikDastaavej}}.

To the best of our knowledge, this is the first work in the Indian legal domain focusing on automated private legal document generation. Our key contributions include:

\begin{enumerate}
    \item \texttt{VidhikDastaavej} Dataset: A novel, anonymized dataset of private legal documents for structured legal text generation.
    \item Model-Agnostic Wrapper: A structured framework ensuring coherence, consistency, and factual accuracy in generated legal drafts.
    \item Expert-Based Evaluation Metrics: Introduction of structured legal evaluation focusing on factual accuracy and completeness.
    \item Human-in-the-Loop System: A user-friendly interface for structured legal document generation, supporting practical legal workflows.
\end{enumerate}

This research lays the foundation for AI-assisted legal drafting in India, modernizing legal workflows while ensuring accuracy, consistency, and legal compliance.

\section{Related Work}
\label{sec:related_work}

AI and NLP have advanced significantly in the legal domain, supporting applications such as judgment prediction~\cite{medvedeva2023legal}, legal case summarization~\cite{ragazzi2024lawsuit, moro2024multi, shukla2022legal}, semantic segmentation~\cite{moro2022semantic}, and legal NER~\cite{puaiș2021named}. These studies demonstrate the potential of AI systems to improve transparency, efficiency, and explainability in legal practice. In the Indian context, prior work has largely focused on public case judgments, particularly from the Supreme Court and High Courts, for retrieval, reasoning, and explainability tasks~\cite{chalkidis2020legal}. Datasets such as ILDC~\cite{malik-etal-2021-ildc},  PredEx~\cite{nigam-etal-2024-legal}, and NyayaAnumana~\cite{nigam-etal-2025-nyayaanumana} support judgment prediction and rationale extraction, while rhetorical role labeling and segmentation tasks have been addressed using models ranging from CRF-BiLSTM~\cite{bhattacharya2019identification} to hierarchical frameworks like HiCuLR~\cite{santosh2024hiculr}. These datasets facilitate training transformer-based models to enhance explainability and decision support systems for Indian legal texts~\cite{nigam2022nigam, malik-etal-2022-semantic, nigam2023legal}, while recent studies have also addressed legal NER using large-scale pretrained models~\cite{vats2023llms}.

Globally, legal text generation has begun to gain traction, extending beyond judgment analysis to structured drafting tasks. Early studies explored controlled natural language drafting~\cite{tateishi2019automatic}, segmentation-assisted contract generation~\cite{tong2022smart}, and text style transfer models~\cite{li2021tst}. Knowledge graph-based approaches have also been proposed to enforce coherence in drafting~\cite{wei2024intelligent}. Recent advances include retrieval-augmented and planning-driven methods for legal drafting and summarization: LexGenie~\cite{t-y-s-s-etal-2025-lexgenie} generates structured multi-case reports for European Court of Human Rights (ECHR) law; CLERC~\cite{hou-etal-2025-clerc} enables case retrieval and retrieval-augmented reasoning for U.S. law; LexKeyPlan~\cite{t-y-s-s-hernandez-2025-lexkeyplan} introduces anticipatory keyphrase planning to improve retrieval-grounded generation; and CoPERLex~\cite{t-y-s-s-etal-2025-coperlex} demonstrates event-centric content planning for case summarization. In the legislative domain, LexDrafter~\cite{chouhan-gertz-2024-lexdrafter} employs RAG for harmonized terminology drafting in EU legal texts. More recently, tools like LEGALSEVA~\cite{pandey2024legalseva} and Legal DocGen~\cite{patil2024legal} demonstrate efforts toward automated drafting of legal contracts and forms.

Despite these global developments, automation of \textit{private} legal document drafting in India remains largely unexplored due to the unavailability of curated datasets. Our work fills this gap by introducing \texttt{VidhikDastaavej}, a large-scale anonymized dataset of diverse private legal documents, and proposing a structured, model-agnostic wrapper framework for coherent and factually accurate document generation. This positions our work at the intersection of dataset creation, structured generation methodology, and rigorous legal expert evaluation.

\section{Problem Statement}

The primary objective of this work is to develop a system that can automatically generate private legal documents based on specific user prompts or situational inputs. Given an input \( x \), which includes detailed instructions or contextual information, the task is to produce a legal document \( y \) that aligns with professional legal drafting standards in the Indian legal domain.

Formally, the problem can be defined as learning a function \( f \) such that: $y = f(x)$,
%
where:

\begin{itemize}
    \item \( x \) represents the user-provided prompt containing specific instructions, situational details, and any particular requirements for the legal document.
    \item \( y \) is the generated legal document that accurately reflects the content of \( x \) and is properly formatted and structured according to legal conventions.
\end{itemize}

The challenge lies in accurately mapping the input \( x \) to a coherent and contextually appropriate document \( y \). This requires the system to understand and interpret complex legal language, terminologies, and document structures specific to the Indian legal context. The goal is to leverage LLMs to perform this mapping effectively, enabling the generation of high-quality legal documents that meet professional standards.

\section{Dataset}
\begin{table}[t]
\centering
\resizebox{0.9\columnwidth}{!}{%
\begin{tabular}{lrr}
\toprule
\textbf{Metric} & \textbf{Train} & \textbf{Test} \\
\midrule
Number of documents & 11,692 & 133 \\
Number of unique categories & 133 & 133 \\
Avg \# of words per document & 5,798.61 & 7,464.62 \\
Max \# of words per document & 98,607 & 81,233 \\
\bottomrule
\end{tabular}
}
\caption{Dataset statistics for \texttt{VidhikDastaavej}.}
\label{table:datastats}
\end{table}
To develop our automated legal document generation tool, we collaborated with an Indian legal firm to curate \texttt{VidhikDastaavej}, a novel, large-scale, anonymized dataset of private legal documents. This partnership granted access to a diverse collection of legal drafts that are not publicly available, ensuring that our dataset reflects real-world legal drafting practices in the Indian legal system. The dataset is designed not only as training material but also as a benchmark for the broader Legal AI community.

\subsection{Dataset Composition and Diversity}
The dataset encompasses a wide variety of documents, including License Agreements, Severance Agreements, Stock Option Agreements, Consulting Agreements, Asset Purchase Agreements, and more. By incorporating multiple document types, \texttt{VidhikDastaavej} captures diverse structures, terminologies, and drafting conventions in legal writing, moving beyond the traditional focus on case judgments seen in public legal datasets.

\begin{figure*}[t]
    \centering
    \includegraphics[width=0.8\linewidth]{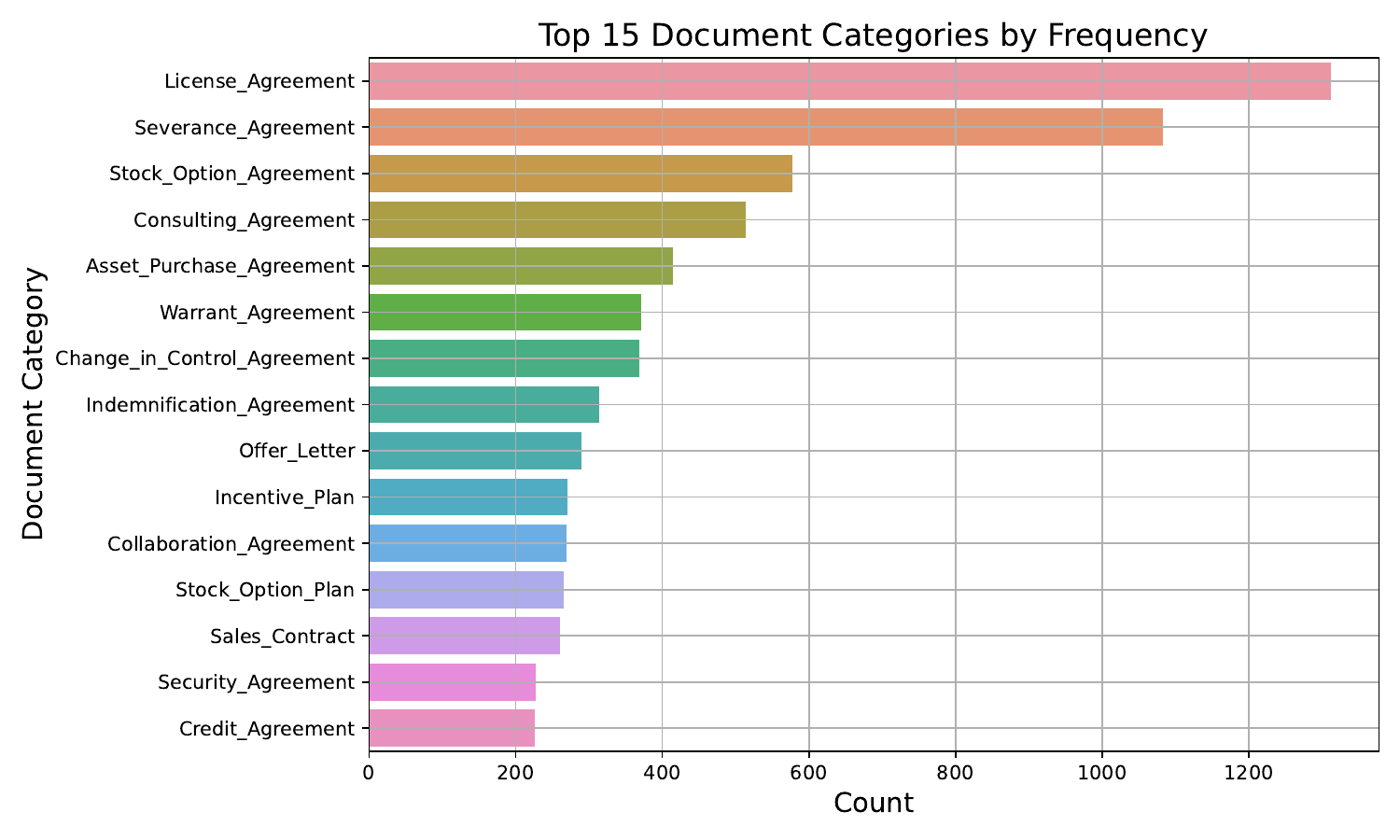}
    \caption{Top 15 Document Categories in the \texttt{VidhikDastaavej} Dataset based on Frequency.}
    \label{fig:top_15_doc_categories}
\end{figure*}

Table~\ref{table:datastats} provides an overview of the dataset statistics. \texttt{VidhikDastaavej} consists of 11,825 documents, with 11,692 used for training and 133 reserved for testing. The dataset covers 133 distinct legal document categories in both training and testing, offering a broad representation of real-world legal drafts. A visual distribution of the top 15 most frequent document categories is presented in Figure~\ref{fig:top_15_doc_categories}. The complete category list is provided in the anonymous GitHub repository and dataset folder.

\subsection{Annotation and Expert Curation}
Since legal drafting requires domain expertise, we employed a multi-stage annotation and validation pipeline:
\begin{itemize}
    \item \textbf{Document Categorization:} Initial categorization was performed using LLM-based classification (Mixtral) followed by validation from practicing legal experts. Each document was mapped to one of 133 categories. Experts verified the accuracy of categorization to avoid propagation of noisy labels.
    \item \textbf{Section-Level Structuring:} For training prompts, section headers were generated using LLaMA-3.1 models and then manually reviewed by experts to ensure alignment with professional drafting norms (e.g., inclusion of clauses, definitions, governing law sections).
    \item \textbf{Expert Instructions:} Experts were provided with clear guidelines on assessing legal soundness. They were asked to evaluate factual accuracy (whether the draft adhered to given instructions and contained no hallucinations) and completeness (coverage of mandatory sections and details). These were rated on a 1--10 Likert scale.
    \item \textbf{Inter-Annotator Reliability:} To ensure reliability, three independent legal experts reviewed the generated drafts, and inter-annotator agreement metrics (Fleiss' $\kappa$, ICC, Krippendorff's $\alpha$) were computed. Results demonstrated strong agreement, validating the robustness of expert evaluations.
\end{itemize}

\subsection{Data De-identification (NER-based Anonymization)}
\label{sec:deid}

To comply with privacy regulations and ethical standards, all documents underwent rigorous anonymization. All documents were de-identified prior to any analysis or model training. We applied a Named Entity Recognition (NER) based redaction procedure using spaCy (\texttt{en\_core\_web\_sm}). For each document, spaCy detects entity spans (e.g., \texttt{PERSON}, \texttt{ORG}, \texttt{GPE}, \texttt{LOC}, \texttt{DATE}) and we replace each detected span with a type placeholder of the form \texttt{[LABEL]}.

Formally, given a document $x$ and the set of detected entity spans
$\mathcal{E}(x)=\{(s_i, e_i, \ell_i)\}_{i=1}^{n}$,
where $(s_i,e_i)$ are character offsets and $\ell_i$ is the entity label, we produce a de-identified document $\tilde{x}$ by replacing every substring $x[s_i:e_i]$ with \texttt{[}$\ell_i$\texttt{]} while preserving all non-entity text unchanged. Table~\ref{table:after_anonymize} shows an example output.

\paragraph{Human verification.}
Because automated NER-based redaction can miss identifiers depending on writing style and context, we conducted expert manual inspections on a subset of the corpus to check for residual personally identifying information. The inspected subset did not reveal remaining identifiers in the reviewed documents; however, we acknowledge that no automated method can guarantee perfect anonymization in all cases.



\subsection{Significance of the Dataset}
Unlike prior datasets that focus primarily on court judgments or narrow subsets of legal texts, \texttt{VidhikDastaavej} provides a broad, real-world representation of private legal documentation in India. This diversity allows models to learn structural conventions, drafting styles, and domain-specific vocabulary. The dataset serves as a foundational benchmark for the Legal AI community, supporting tasks such as document generation, section planning, and factual verification. By releasing both the anonymized corpus and the evaluation setup, we aim to foster reproducible research and enable further advancements in AI-assisted legal drafting.

\begin{table*}[t]
\centering
\resizebox{0.8\linewidth}{!}{%
\begin{tabular}{|p{15cm}|}
\hline
\textbf{Power of Attorney} \newline
\textit{To All of whom, these presents shall come, I [PERSON] of [GPE] send Greetings} \newline
\textbf{Whereas,}
\begin{enumerate}
    \item Mr. [PERSON] shall appoint some fit and proper person to carry on acts for me and manage all my affairs.
    \item I nominate, constitute, and appoint my brother, Mr. [PERSON], as my true and lawfully appointed attorney (hereinafter called the Attorney) to act for me in the court of law for court proceedings in the matter of disputed joint property.
\end{enumerate}
{NOW THIS PRESENT WITNESSETH AS FOLLOWS:}
\begin{enumerate}
    \item  The attorney shall handle all the affairs with regard to court proceedings in the matter of disputed joint property.
    \item  All the filings of applicants and suits in the court of law.
    \item  All the appearances in the court proceedings.
    \item  All the costs, expenses, and fees with regard to court proceedings.
    \item  The fees to be paid to the lawyer appointed.
\end{enumerate}
\textit{And I, Mr. [PERSON], undertake to ratify all the acts of the attorney or any agent appointed by him.} \newline
{IN WITNESS WHEREOF}, I set and subscribe my hand on [DATE]. \newline
\_\_\_\_\_\_\_\_\_\_ \newline
\textit{[WORK\_OF\_ART] by within named.} \newline
Mr. [PERSON] above named in the presence of: \newline
\begin{enumerate}
    \item  \_\_\_\_\_\_\_\_\_\_ Mr. [PERSON] \newline
    \item  \_\_\_\_\_\_\_\_\_\_ Mr. [PERSON]
\end{enumerate} \\
\hline
\end{tabular}
}
\caption{This table illustrates a sample document after it has been anonymized.}
\label{table:after_anonymize}
\end{table*}

\section{Model-Agnostic Wrapper (MAW)}
\begin{figure*}[t]
\centering
\includegraphics[width=\linewidth]{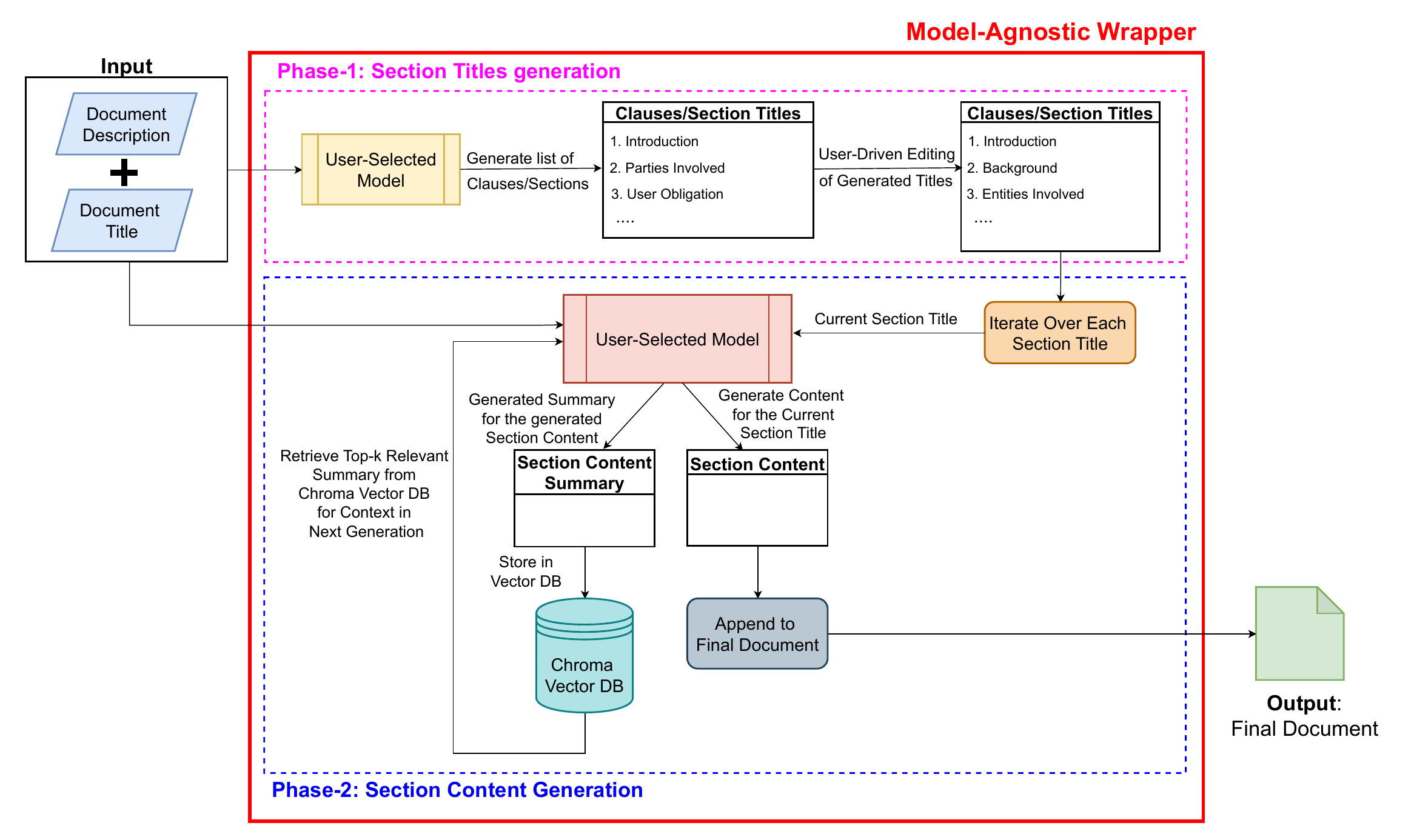}
\caption{Wrapper flow diagram}
\label{fig:Wrapper-Flow-Diagram}
\end{figure*}

To improve long-form legal document generation, we introduce a Model-Agnostic Wrapper (MAW), a framework designed to integrate with any LLM for structured drafting. Legal documents require maintaining logical flow, coherence, and factual accuracy, which general-purpose LLMs often struggle with when handling extended text generation. 

The MAW employs a two-phase workflow (Figure~\ref{fig:Wrapper-Flow-Diagram}) to ensure structured, contextually relevant content generation.

\noindent
\textbf{Phase 1: Section Title Generation.}  
In the first phase, section titles are generated based on user input. The process begins with the user providing a document title and a brief description of the intended document. These inputs are passed to the chosen language model, which then generates a structured list of section titles. The generated section titles are displayed to the user, who can review and modify them, renaming, inserting new sections, or removing unnecessary ones before proceeding to content generation. Once the section titles are finalized, the process transitions to the next phase.

\noindent
\textbf{Phase 2: Section Content Generation.}  
In the second phase, content is generated iteratively for each section. The workflow follows these steps:
\begin{enumerate}
    \item For each section title, the model receives document title and description as additional context.
    \item The model generates detailed section content along with a concise summary of the section.
    \item The generated summary is stored in a vector database (ChromaDB)~\cite{chromadb2023} to facilitate contextual referencing.
    \item During subsequent iterations, the vector database is queried for relevant section summaries, which are then incorporated into the LLM’s context to enhance coherence and maintain logical document flow.
    \item After generating content for all sections, the final document is refined and structured, ensuring clarity and coherence.
\end{enumerate}

By adopting a two-phase workflow, we ensure that adequate time is dedicated to both section title generation and section content generation separately, rather than attempting to generate both simultaneously. This separation allows for better coherence, logical structuring, and improved alignment between titles and their corresponding content, thereby enhancing the overall quality and readability of the generated document.

To demonstrate practical usability, we have developed a Human-in-the-Loop (HITL) Document Generation System that allows users to specify document types, refine sections, and interactively generate drafts. Due to LREC submission policies, the complete user guide with detailed instructions and screenshots is not included at this stage, but will be provided in the camera-ready version as supplementary material.

\section{Experimental Setup}
\label{sec:experimental_setup}
To benchmark our pipeline's performance and assess our wrapper's effectiveness, we conducted instruction tuning on various open-source models and compared them against GPT-4o. 

\subsection{Fine Tuning of Open-Source Models}
We fine-tuned select open-source models while directly evaluating others without additional training. The instruction-tuned models include Qwen3-14B~\cite{yang2025qwen3}, LLaMA-3.1-8B-Instruct~\cite{dubey2024llama}, and Gemma-3-12B-It~\cite{team2025gemma} SFT to assess improvements in structured legal drafting.

For instruction tuning, we designed specialized prompts and instruction sets tailored to legal drafting. These instructions provided structured examples, ensuring that the models understood the nuances of different types of legal documents. 

\subsection{Benchmarking with GPT-4o}
To assess the effectiveness of our instruction-tuned models and the Model-Agnostic Wrapper, we benchmarked performance against GPT-4o, a proprietary closed-source model. Unlike the open-source models, GPT-4o was not instruction-tuned but was used purely for inference. This comparison highlights the potential of fine-tuned open-source models as cost-effective alternatives for structured legal drafting, offering insights into whether instruction tuning can achieve performance comparable to commercial LLMs.

\begin{table}[t]
\centering
\resizebox{0.5\textwidth}{!}{%
\begin{tabular}{|p{2cm}|p{14cm}|}
\hline
\multicolumn{1}{|l|}{\textbf{Category}} & \multicolumn{1}{c|}{\textbf{Prompt}} \\
\hline

\textbf{Development Agreement} & 
Create a development, license, and hosting agreement between [ORG] and [ORG] LLC, effective as of [DATE], outlining the terms and conditions for the development, licensing, and hosting of [ORG], including [ORG] and [ORG], for the sale of [ORG] flights and other air transportation services through the [ORG] website. The agreement should include provisions for the definition of key terms, the scope of work, the schedule, the fees, the payment terms, the confidentiality obligations, the intellectual property rights, the warranties and disclaimers, the indemnification obligations, the limitation of liability, the insurance requirements, the dispute resolution procedures, the term and termination provisions, and the miscellaneous provisions. The agreement should also include exhibits for the specifications, the change request, the schedule, the fees, the relationship managers, the service level agreement, the non-disclosure agreement, and the escrow agreement \\
\hline

\textbf{Purchase Agreement} & 
Create a purchase agreement between [WORK OF ART] and Stacked Digital LLC, outlining the terms and conditions of the sale, including the purchase price, payment terms, delivery schedule, warranties, and any other relevant details necessary for a comprehensive agreement between the [CARDINAL] parties. \\
\hline

\end{tabular}%
}
\caption{Categories and Corresponding Prompts for Legal Document Generation}
\label{table:legal_prompts}
\end{table}

\subsection{Hyperparameters}
\label{appendix:experimental_setup}
All experiments were conducted using the PyTorch framework integrated with Hugging Face Transformers. For SFT (Supervised Fine-Tuning), we used four NVIDIA H200 (Neysa) GPUs with 80GB of memory each. Mixed-precision training (fp16) was enabled to optimize memory and computational efficiency, and training progress was logged with Weights \& Biases for effective monitoring.

We fine-tuned three instruction models, Qwen3-14B, Gemma-3-12B-It, and LLaMA-3.1-8B-Instruct, on the expanded dataset. Each model supported a maximum sequence length of 4500 tokens, allowing for long-context learning essential for legal drafting tasks.

The optimization was performed using the AdamW optimizer with a learning rate of \(1 \times 10^{-4}\), paired with a cosine learning rate scheduler for stable decay. We employed gradient accumulation over 4 steps (per-device batch size: 1, effective batch size: 4) and trained all models for 3 epochs. These settings provided a balance between performance and training resource constraints.

To guide the models during SFT, we prepared a diverse set of instruction prompts that encapsulated real-world legal drafting scenarios, ensuring relevance and structure. Sample prompts are shown in Table~\ref{table:legal_prompts}, and the complete set will be made public after acceptance to support reproducibility and further research in legal document generation.

\section{Evaluation Metrics}
To evaluate model performance in legal document generation, we adopt a multi-dimensional approach combining automatic metrics and expert judgments. This ensures coverage of surface-level similarity, semantic quality, coherence, and legal soundness.

Lexical accuracy was measured using ROUGE-L~\cite{lin-2004-rouge}, BLEU~\cite{papineni-etal-2002-bleu}, and METEOR~\cite{banerjee-lavie-2005-meteor}, which assess overlap between generated and reference texts. Semantic quality was assessed using BERTScore~\cite{BERTScore} and BLANC~\cite{blanc}, which capture contextual and meaning-level alignment.

To evaluate beyond similarity, we employed G-Eval~\cite{liu-etal-2023-g}, a GPT-4-based framework for assessing factuality, coherence, and completeness. This allowed us to quantify performance on structured reasoning dimensions critical for legal texts.

Given the domain-specific nature of drafting, human expert evaluation was indispensable. Three legal professionals assessed each generated draft as a whole, rating them on two criteria: (i) Factual Accuracy- adherence to given instructions and legal facts, and (ii) Completeness \& Comprehensiveness- coverage of all necessary legal elements. Since the target audience is ultimately legal experts, this step ensured that evaluations reflected real-world drafting needs.

To confirm reliability, we computed inter-annotator agreement (IAA) across expert ratings using Fleiss' $\kappa$~\cite{fleiss1971measuring}, Cohen's $\kappa$~\cite{cohen1960coefficient}, Intraclass Correlation Coefficient (ICC)~\cite{shrout1979intraclass}, Krippendorff's $\alpha$~\cite{krippendorff2018content}, and Pearson correlation~\cite{benesty2009pearson}. Higher agreement was observed for structured wrapper outputs, showing that section-wise generation aids consistent expert judgments.

\begin{table}[ht]
\centering
\resizebox{\columnwidth}{!}{%
\begin{tabular}{|l|}
\hline
\textbf{Instructions}:\\
You are an expert in legal text evaluation. You will be given:\\

A document description that specifies the intended content of a generated legal document.\\
An actual legal document that serves as the reference. A generated legal document that\\ needs to be evaluated.
Your task is to assess how well\ the generated document aligns with\\ the given description while using the actual document as a reference for correctness.\\
\\

\textbf{Evaluation Criteria (Unified Score: 1-10)}\\
Your evaluation should be based on the following factors:\\

\textit{Factual Accuracy (50\%)} – Does the generated document correctly represent the key legal\\ facts, reasoning, and outcomes from the original document, as expected from the description?\\
\textit{Completeness \& Coverage (30\%)} – Does it include all crucial legal arguments, case details,\\ and necessary context that the description implies?\\
\textit{Clarity \& Coherence (20\%)} – Is the document well-structured, logically presented,\\ and legally sound?\\
\\
\textbf{Scoring Scale:}\\
1-3 → Highly inaccurate, major omissions or distortions, poorly structured.\\
4-6 → Somewhat accurate but incomplete, missing key legal reasoning or context.\\
7-9 → Mostly accurate, well-structured, with minor omissions or inconsistencies.\\
10 → Fully aligned with the description, factually accurate, complete, and coherent.\\
\\
\textbf{Input Format:}\\
Document Description:\\
\{\{doc\_des\}\}\\
\\
\textbf{Original Legal Document (Reference):}\\
\{\{Actual\_Document\}\}\\
\\
\textbf{Generated Legal Document (To Be Evaluated):}\\
\{\{Generated\_Document\}\}\\
\\
\textbf{Output Format:}\\
Strictly provide only a single integer score (1-10) as the response,\\with no explanations, comments, or additional text.\\
\hline
\end{tabular}}
\caption{The prompt is utilized to obtain scores from the G-Eval automatic evaluation methodology. We employed the GPT-4o-mini model to evaluate the quality of the generated text based on the provided prompt/input description, alongside the actual document as a reference.}
\label{tab:G-Eval Prompt}
\end{table}

\section{Results and Analysis}

\begin{table*}[t]
\centering
\resizebox{0.8\linewidth}{!}{%
\begin{tabular}{lccccccccc}
\toprule
\textbf{Models} & \multicolumn{3}{c}{\textbf{Lexical Based Evaluation}} & \multicolumn{2}{c}{\textbf{Semantic Evaluation}} & \textbf{Automatic LLM} & \multicolumn{2}{c}{\textbf{Average Expert Scores}} \\
\cmidrule(lr){2-4} \cmidrule(lr){5-6} \cmidrule(lr){7-7} \cmidrule(lr){8-9} & RL & BLEU & METEOR & BERTScore & BLANC & G-Eval & Factual & Completeness \& \\ 
& & & & & & & Accuracy & Compre. \\
\midrule
Qwen3-14B & 0.09 & 0.00 & 0.07 & 0.73 & 0.01 & 3.56 & 1.00 & 1.00 \\
LLaMA-3.1-8B-Instruct &  0.09 & 0.01 & 0.11 & 0.78 & 0.04 & 1.57 & 1.00 & 1.10 \\
LLaMA-3.1-8B-Instruct SFT & 0.08 & 0.00 & 0.05 & 0.74 & 0.01 & 1.12 & 1.00 & 1.00 \\
Wrapper (Over LLaMA-3.1-8B) &  \textbf{0.15} & 0.04 & 0.18 & 0.79 & 0.19 & 5.15 & 3.30 & 2.20 \\
Gemma-3-12B-It & 0.09 & 0.01 & 0.10 & 0.76 & 0.02 & 1.13 & 1.00 & 1.00 \\
Gemma-3-12B-It SFT & 0.11 & 0.01 & 0.10 & 0.78 & 0.04 & 1.37 & 1.00 & 1.00 \\
Wrapper (Over Gemma-3-12B) & \textbf{0.15} & \textbf{0.06} & \textbf{0.24} & 0.80 & 0.17 & 6.56 & \textbf{8.82} & \textbf{7.82} \\
GPT-4o & 0.14 & 0.03 & 0.12 & \textbf{0.81} & \textbf{0.24} & \textbf{6.68} & 8.80 & 5.40 \\
\bottomrule
\end{tabular}}
\caption{Evaluation metrics for new models. LLaMA-3.1-8B-Instruct and Gemma-3-12B denote the instruction-tuned variants of their respective base models. The best scores are highlighted in bold.}
\label{table:model_evaluation}
\end{table*}

This section presents the evaluation results of various models for legal document generation. The models were assessed using lexical-based, semantic similarity-based, automatic LLM-based, and expert evaluation metrics, as detailed in Table~\ref{table:model_evaluation}. Our findings highlight key challenges, the impact of supervised fine-tuning (SFT), and the effectiveness of the model-agnostic wrapper.

\subsection{Comparative Model Performance}
Among the open-source models, Qwen3-14B, LLaMA-3.1-8B-Instruct, and Gemma-3-12B-It exhibited limited performance across both lexical and semantic evaluations. More importantly, direct supervised fine-tuning (SFT) on these models led to further performance degradation. This result, while initially surprising, can be explained by several underlying factors that we analyzed in depth.  
One major factor was dataset diversity and imbalance. Even though the \texttt{VidhikDastaavej} dataset was expanded to more than 11,000 documents across 133 categories, certain specialized legal document types remained underrepresented. As a result, SFT models tended to overfit to the dominant patterns in frequently occurring categories while struggling to generalize to underrepresented ones. This behavior was evident in outputs where SFT models reproduced overly generic clauses or omitted critical sections in less frequent document types.  

Another factor was the mismatch between the instruction style used in SFT training and the structured prompting strategy employed in our wrapper. SFT relied on flat, single-shot instructions paired with full document targets, which constrained the model to a rigid learning signal. In contrast, the wrapper decomposed document generation into a two-phase workflow: planning section titles and generating section-wise content with retrieval augmentation. This structured and iterative approach aligns more closely with how legal professionals draft documents, ensuring global coherence and better factual consistency.  

A third important consideration is the role of retrieval in grounding. While SFT attempted to internalize drafting knowledge directly from limited training samples, the wrapper dynamically retrieved relevant context for each section at generation time. This retrieval-augmented process reduced hallucinations, ensured clause-specific accuracy, and allowed even smaller base models to produce outputs that were coherent and factually grounded. In practice, this adaptability compensated for data sparsity and improved performance across evaluation metrics.  

As shown in Table~\ref{table:model_evaluation}, wrapper-enhanced models consistently outperformed both base and SFT variants. For instance, Gemma-3-12B-It achieved an expert factual accuracy score of only 1.00 after SFT, whereas the wrapper applied over the same model yielded a score of 8.82. Similarly, LLaMA-3.1-8B-Instruct showed a marked jump in completeness when used with the wrapper. These results demonstrate that retrieval-based, modular prompting is more effective than conventional fine-tuning for long-form, high-precision domains such as legal drafting.  

Finally, a closer inspection of failure cases highlighted common shortcomings of SFT models. For example, in generating a Shareholders’ Agreement, SFT outputs frequently omitted key provisions such as ``Governing Law'' or ``Termination Clauses,'' or invented irrelevant citations. Wrapper-enhanced models, by contrast, generated structurally complete drafts with higher factual fidelity, aided by their explicit planning and retrieval steps. Illustrative hallucination examples are included in Table~\ref{tab:hallucination_example}.  

Overall, these findings suggest that while SFT remains a valuable adaptation technique, its effectiveness is limited in domains like law where diversity, structure, and factual precision are paramount. The model-agnostic wrapper provides a more reliable and scalable solution, ensuring consistency across models and enabling practical deployment in real-world legal drafting tasks.

\begin{table*}[t]
\centering
\resizebox{0.8\linewidth}{!}{%
\begin{tabular}{|p{0.25\textwidth}|p{0.40\textwidth}|p{0.40\textwidth}|}
\hline
\textbf{Prompt} & \textbf{Reference Output (Correct Draft)} & \textbf{Generated Output (Hallucinated)} \\
\hline
\begin{minipage}[t]{\linewidth}
Mr. [PERSON], an elder brother, wants to authorize his brother Mr. [PERSON] by giving power of attorney to appear in the court of law for court proceedings in the matter of disputed joint property. Draft a power of attorney.
\end{minipage}
&
\begin{minipage}[t]{\linewidth}
\textbf{Power of Attorney} \\
I, Mr. [PERSON], hereby appoint my brother, Mr. [PERSON], to act on my behalf in all legal proceedings concerning the disputed joint property contested by our relatives. He shall represent me in court, file applications, respond to notices, and bear necessary expenses. \\
IN WITNESS WHEREOF, I sign this on [DATE].
\end{minipage}
&
\begin{minipage}[t]{\linewidth}
\textbf{General Power of Attorney} \\
I, Mr. [PERSON], appoint Mr. [PERSON] to manage all my financial and real estate matters, including buying, selling, and mortgaging properties across India. He may also represent me in taxation and banking disputes. \\
IN WITNESS WHEREOF, I grant him full authority without limitation. 
\end{minipage}
\\
\hline
\begin{minipage}[t]{\linewidth}
Draft a Lease Agreement between Mr. [PERSON] (landlord) and Ms. [PERSON] (tenant) for a residential flat in [LOCATION] for 11 months at a monthly rent of 15,000 INR.
\end{minipage}
&
\begin{minipage}[t]{\linewidth}
\textbf{Lease Agreement} \\
This agreement is made between Mr. [PERSON] (landlord) and Ms. [PERSON] (tenant) for a residential flat at [LOCATION]. The term of lease shall be 11 months commencing on [DATE], with a monthly rent of 15,000 INR. The tenant agrees not to sublet the premises. Governing law: Indian Contract Act. 
\end{minipage}
&
\begin{minipage}[t]{\linewidth}
\textbf{Lease Agreement} \\
This agreement is made between Mr. [PERSON] (landlord) and Ms. [PERSON] (tenant) for commercial office space at [LOCATION]. The term of lease shall be 5 years, with a monthly rent of 25,000 INR. The tenant may sublet the premises with prior written notice. Governing law: English Law.
\end{minipage}
\\
\hline
\end{tabular}
}
\caption{Examples of hallucinations in AI-generated legal drafting. Unlike incoherent outputs, hallucinations manifest as factual inconsistencies (e.g., inventing clauses, altering contract type, or changing law).}
\label{tab:hallucination_example}
\end{table*}

\subsection{Effectiveness of MAW}
One of the most promising findings of our study is the effectiveness of the model-agnostic wrapper in generating structured, large, and coherent legal documents. The wrapper enhances consistency across sections, ensuring logical flow and improving document quality. This method proves particularly effective for maintaining coherence in complex legal texts, overcoming the limitations of individual models. Notably, the wrapper’s outputs achieved comparable scores to GPT-4o, despite being generated using open-source models. Expert evaluations further confirm that the generated documents from wrapper-assisted models were coherent, well-structured, and legally valid, demonstrating the utility of this approach.

An additional advantage of the wrapper function is its ability to reduce hallucinations in legal text generation. Hallucinations, where the model generates factually incorrect or legally inconsistent information, pose a significant challenge in AI-generated legal documents. By enforcing a structured, stepwise document generation approach, the wrapper minimizes it by ensuring that the generated content remains grounded in the given instructions and previously generated sections.

\subsection{Expert Evaluation: Factual Accuracy and Completeness}
Expert evaluation provides the most reliable measure of an AI-generated document’s real-world applicability. Our findings show that factual accuracy and completeness scores correlate strongly with expert assessments, highlighting their importance as legal-specific evaluation metrics. Models that underwent SFT struggled with maintaining factual consistency, likely due to the limited amount of the fine-tuning dataset. On the other hand, the MAW significantly improved both factual accuracy and completeness, reinforcing its role in enhancing document consistency and legal validity. Wrapper-enhanced models received high marks, with the Gemma-based wrapper achieving expert ratings of 8.82 (factual) and 7.82 (completeness), ahead of GPT-4o. This suggests wrapper-based prompting can offer performance comparable to proprietary models in specialized domains like legal NLP.  

\subsection{IAA Findings and Observations}

Tables~\ref{table:iaa_factual_accuracy} and~\ref{table:iaa_completeness} summarize the IAA results for the factual accuracy and completeness scores, respectively. We observed moderate agreement for baseline models such as Qwen3-14B and LLaMA-3.1-8B-Instruct. However, the wrapper-enhanced configurations exhibited consistently higher agreement scores, indicating their outputs were easier for experts to evaluate consistently.

Wrapper-based variants achieved Fleiss’ $\kappa$ and Krippendorff’s $\alpha$ above 0.80, and ICC values approaching or exceeding 0.90 for factual accuracy, highlighting strong consensus among raters. Completeness scores showed similar trends, reinforcing that structured generation enhances clarity and assessment consistency.
GPT-4o also demonstrated high agreement, but the best-performing wrapper-based open-source models were competitive, validating their utility as viable alternatives.

\subsection{Insights from Legal Experts}
\label{sec:legal_expert_insights}

In addition to numerical scoring, legal experts provided detailed qualitative feedback. Key insights:

\begin{itemize}
    \item \textit{Improved Structure and Coherence}: Experts appreciated that wrapper-based outputs exhibited logical progression and better adherence to legal formatting norms, particularly in section-wise organization.
    \item \textit{Reduced Hallucinations}: Experts found outputs from wrapper-based models to be more factually grounded, supported by improved use of domain-specific terminology and reduced irrelevancy.
    \item \textit{Linguistic Clarity and Formalism}: Continued pretraining was noted to improve the quality of formal legal language. Experts preferred drafts mimicking Indian legal writing conventions.
    \item \textit{Areas for Improvement}: Minor verbosity and occasional factual inconsistencies were observed in longer drafts. Experts recommended integrating case precedents and statutory references for more robust legal drafting.
\end{itemize}


\begin{table}[t]
\centering
\resizebox{\linewidth}{!}{%
\begin{tabular}{lccccc}
\toprule
\textbf{Models} & \textbf{Fleiss' $\kappa$} & \textbf{Cohen's $\kappa$} & \textbf{ICC} & \textbf{Kripp. $\alpha$} & \textbf{Pearson Corr.} \\
\midrule
Qwen3-14B & 0.42 & 0.44 & 0.49 & 0.45 & 0.43 \\
Llama-3.1-8B-Instruct & 0.38 & 0.40 & 0.42 & 0.41 & 0.39 \\
Llama-3.1-8B-Instruct SFT & 0.35 & 0.39 & 0.41 & 0.39 & 0.37 \\
Wrapper Over (Llama-3.1-8B) & 0.79 & 0.75 & 0.88 & 0.86 & 0.89 \\
Gemma-3-12b-it & 0.33 & 0.38 & 0.39 & 0.38 & 0.35 \\
Gemma-3-12b-it SFT & 0.30 & 0.36 & 0.35 & 0.36 & 0.33 \\
Wrapper Over (Gemma-3-12B) & 0.81 & 0.80 & 0.91 & 0.90 & 0.92 \\
GPT4o & 0.82 & 0.84 & 0.89 & 0.87 & 0.91 \\
\bottomrule
\end{tabular}
}
\caption{Inter-Annotator Agreement (IAA) Metrics for Factual Accuracy, evaluating consistency among expert reviewers across different models.}
\label{table:iaa_factual_accuracy}
\end{table}

\begin{table}[t]
\centering
\resizebox{\linewidth}{!}{%
\begin{tabular}{lccccc}
\toprule
\textbf{Models} & \textbf{Fleiss' $\kappa$} & \textbf{Cohen's $\kappa$} & \textbf{ICC} & \textbf{Kripp. $\alpha$} & \textbf{Pearson Corr.} \\
\midrule
Qwen3-14B & 0.40 & 0.42 & 0.48 & 0.44 & 0.42 \\
Llama-3.1-8B-Instruct & 0.37 & 0.39 & 0.41 & 0.39 & 0.38 \\
Llama-3.1-8B-Instruct SFT & 0.36 & 0.38 & 0.40 & 0.38 & 0.36 \\
Wrapper Over (Llama-3.1-8B) & 0.73 & 0.70 & 0.85 & 0.83 & 0.87 \\
Gemma-3-12b-it & 0.34 & 0.37 & 0.37 & 0.36 & 0.34 \\
Gemma-3-12b-it SFT & 0.32 & 0.35 & 0.33 & 0.34 & 0.31 \\
Wrapper Over (Gemma-3-12B) & 0.77 & 0.75 & 0.87 & 0.86 & 0.89 \\
GPT4o & 0.78 & 0.79 & 0.88 & 0.86 & 0.90 \\
\bottomrule
\end{tabular}
}
\caption{IAA Metrics for Completeness \& Comprehensiveness, evaluating consistency among expert reviewers across different models.}
\label{table:iaa_completeness}
\end{table}

\section{Ablation Study}
\subsection{Ablation: Impact of Retrieval Module}
To quantify the contribution of the retrieval module in the Model-Agnostic Wrapper (MAW), we conducted an ablation where retrieval was disabled while keeping the rest of the structured generation pipeline intact. Removing retrieval resulted in a decline in both expert-assessed factual accuracy ($-2.2$ points) and completeness ($-1.7$ points). Lexical metrics and semantic metrics also dropped consistently. This indicates that retrieval plays a crucial role in grounding the generation process with relevant precedents, improving both legal accuracy and contextual alignment.

\subsection{Component-wise Ablation of the Wrapper}
To disentangle the impact of individual components in the wrapper, we evaluated different configurations:
\begin{itemize}
    \item \textit{Long Prompt Only (no structure or retrieval):} Minor improvements in lexical overlap but no noticeable change in factual accuracy.

    \item \textit{Retrieval Only (flat prompt without structure):} Showed moderate gains in completeness but lacked logical flow and legal coherence.

    \item \textit{Structured Generation Only (no retrieval):} Provided better document organization but failed to anchor content in precedent-specific context.
\end{itemize}

Only the full wrapper, combining structured generation and retrieval, consistently achieved high scores across all metrics, most notably $+4.5$ in factual accuracy and $+3.8$ in completeness (expert Likert scores). These findings confirm that both structured planning and contextual grounding are essential to improving legal document generation.

\section{Conclusion and Future Work}
This work introduces VidhikDastaavej, a novel large-scale dataset of private Indian legal documents, and a Model-Agnostic Wrapper (MAW) that enables structured, section-wise legal drafting across models. Our results show that while supervised fine-tuning often degraded performance, the wrapper approach consistently improved coherence, factual accuracy, and completeness, even surpassing strong proprietary models in expert evaluations.
To ensure practical use, we also developed a Human-in-the-Loop system that allows experts to refine and validate drafts interactively. Together, the dataset, wrapper methodology, and expert-based evaluation establish a foundation for robust AI-assisted legal drafting in resource-constrained and evolving model settings.

Future work will expand the dataset with more documents per category, adopt rubric-style domain-specific evaluations to assess legal soundness more rigorously, and integrate factual verification modules and efficiency optimizations. These steps will enhance the scalability, reliability, and real-world usability of AI-driven legal drafting systems.

\section{Acknowledgements}

We would like to express our sincere gratitude to the anonymous reviewers for their constructive feedback and valuable suggestions, which greatly improved the quality of this paper. We also thank the student research assistants and legal experts from various law colleges for their dedicated efforts in annotation and expert evaluation. Their contributions were instrumental in ensuring the quality and reliability of the dataset and evaluation process.
We gratefully acknowledge the support of BharatGen, India, for providing access to computational resources and hardware infrastructure used in this research. Their assistance played a vital role in model training and large-scale experimentation. The majority of this work was conducted while the first author was affiliated with the Indian Institute of Technology Kanpur. The author is currently affiliated with the University of Birmingham Dubai.

\section{Limitations}
Despite the promising results, some limitations remain that define directions for improvement.

First, while the \texttt{VidhikDastaavej} dataset includes over 11,000 documents across 133 categories, the test set includes only one document per category (133 total). This breadth ensures coverage across diverse document types but risks higher variance and possible sensitivity to idiosyncratic cases. To mitigate this, we relied on three independent legal experts and strong inter-annotator agreement metrics (Fleiss' $\kappa$, ICC, Krippendorff's $\alpha$), ensuring reliability beyond raw averages. Nonetheless, expanding test coverage (e.g., $\geq$10 documents per category) will allow for significance testing and deeper per-type analyses.

Second, although the wrapper reduces hallucinations and factual errors, occasional inconsistencies remain. Automated evaluation methods such as G-Eval provide useful signals but can appear ad hoc. A more dedicated evaluation framework—for example, a checklist-style rubric tailored to each document type (mandatory clauses, cross-reference consistency, signature blocks, governing law, etc.), would better capture legal soundness and pinpoint error modes. This remains a key area for future refinement.

Third, the wrapper introduces moderate computational overhead. Compared to direct prompting, we observed roughly a 1.4–1.6× increase in inference time, with retrieval adding 60–80 ms per query. While these costs are justified by substantial quality gains, further optimizations (e.g., caching, adaptive retrieval, and prompt compression) will improve efficiency, particularly in real-world deployments where latency matters.

Finally, although evaluations involved experienced legal experts, broader deployment in professional workflows has not yet been tested. Real-world usage studies will be necessary to validate usability, trust, and adaptability across jurisdictions.

By expanding test coverage, refining domain-specific evaluation protocols, and optimizing computational efficiency, future iterations of this work can deliver even stronger, legally sound, and practically deployable AI solutions for legal drafting.

\section{Ethics Statement}
\label{sec:ethics}

This research studies AI-assisted generation of legal documents, a high-stakes setting with risks related to privacy/confidentiality, bias, transparency, accountability, and potential misuse.
Given the sensitive nature of legal documents, we prioritized data privacy and security in every phase of this study.

\paragraph{Data governance and permissions.}
The dataset \texttt{VidhikDastaavej} was curated in collaboration with a legal firm and used with appropriate permissions for research purposes, and no confidentiality agreements were violated during data collection and use.

\paragraph{Privacy and confidentiality.}
We de-identified documents prior to any processing (technical details in Section~\ref{sec:deid}). Since de-identification may not fully eliminate re-identification risk in all circumstances, we treat the dataset as sensitive and limit access and use accordingly.

\paragraph{Bias and fairness.}
AI models, may inherit biases from historical legal texts, potentially affecting fairness in document generation. To reduce the risk of generating harmful or misleading content, we include expert evaluation criteria focused on factual accuracy and legal completeness, and we position model outputs as drafts requiring professional review.

\paragraph{Transparency, accountability, and human oversight.}
Transparency is crucial in legal AI applications. To improve the reliability of generated documents, we developed the Model-Agnostic Wrapper (MAW), which enforces structured text generation while minimizing hallucinations. However, AI-generated legal drafts are not substitutes for human expertise. The system is designed as an assistive tool, with a Human-in-the-Loop (HITL) mechanism that ensures legal professionals oversee and refine the generated drafts before any official use.

\newpage
\section{Bibliographical References} 
\bibliographystyle{lrec2026-natbib}
\bibliography{anthology,custom}




\end{document}